\documentclass{article}

 \PassOptionsToPackage{numbers, compress}{natbib}
\usepackage[preprint]{neurips_2026}

\usepackage[utf8]{inputenc} % allow utf-8 input
\usepackage[T1]{fontenc}    % use 8-bit T1 fonts
\usepackage{hyperref}       % hyperlinks
\usepackage{url}            % simple URL typesetting
\usepackage{booktabs}       % professional-quality tables
\usepackage{amsfonts}       % blackboard math symbols
\usepackage{nicefrac}       % compact symbols for 1/2, etc.
\usepackage{microtype}      % microtypography
\usepackage{xcolor}         % colors
\usepackage{graphicx}
\usepackage{booktabs}
\usepackage{graphicx}
\usepackage{pifont}
\usepackage{amsmath}
\newcommand{\cmark}{\ding{51}}
\newcommand{\xmark}{\ding{55}}
\usepackage{subcaption}
\title{CoCoSI: Collaborative Cognitive Map Construction for Spatial Intelligence}

\author{ Yiming Zhang$^{2}$ \quad Ruoxuan Cao$^{2}$ \quad Zhihang Zhong$^{1}$\thanks{Corresponding author: \href{mailto:zhihangzhong@sjtu.edu.cn}{\texttt{zhihangzhong@sjtu.edu.cn}}} \\ $^{1}$Shanghai Jiao Tong University \quad $^{2}$Cornell University }
\begin{document}

\maketitle

\begin{abstract}

Spatial intelligence is a key frontier for multimodal large language models (MLLMs), enabling them to reason about the physical world from visual experience. Inspired by human spatial cognition, recent approaches construct grid-based cognitive maps from multi-frame visual inputs to maintain coherent spatial representations over time. However, limited context lengths still challenge spatial understanding, while existing methods, such as long-context modeling and external memory, often require architectural changes, memory modules, or finetuning, limiting their applicability to off-the-shelf pretrained MLLMs. This motivates a lightweight, model-agnostic method for preserving spatial information beyond the native context window. To this end, we propose a plug-and-play multi-agent framework that collaboratively constructs cognitive maps as structured spatial memory, enhancing the spatial understanding of arbitrary pretrained MLLMs without architectural modification or additional training. Our framework features local-global agent coordination, cognitive map construction with atomic commits, and cross-agent verification. Extensive experiments demonstrate that our method achieves superior performance on spatial understanding tasks while remaining fully training-free. Code will be released.
\end{abstract}

\section{Introduction}

\begin{figure}[h!]
    \centering
    \includegraphics[width=\linewidth]{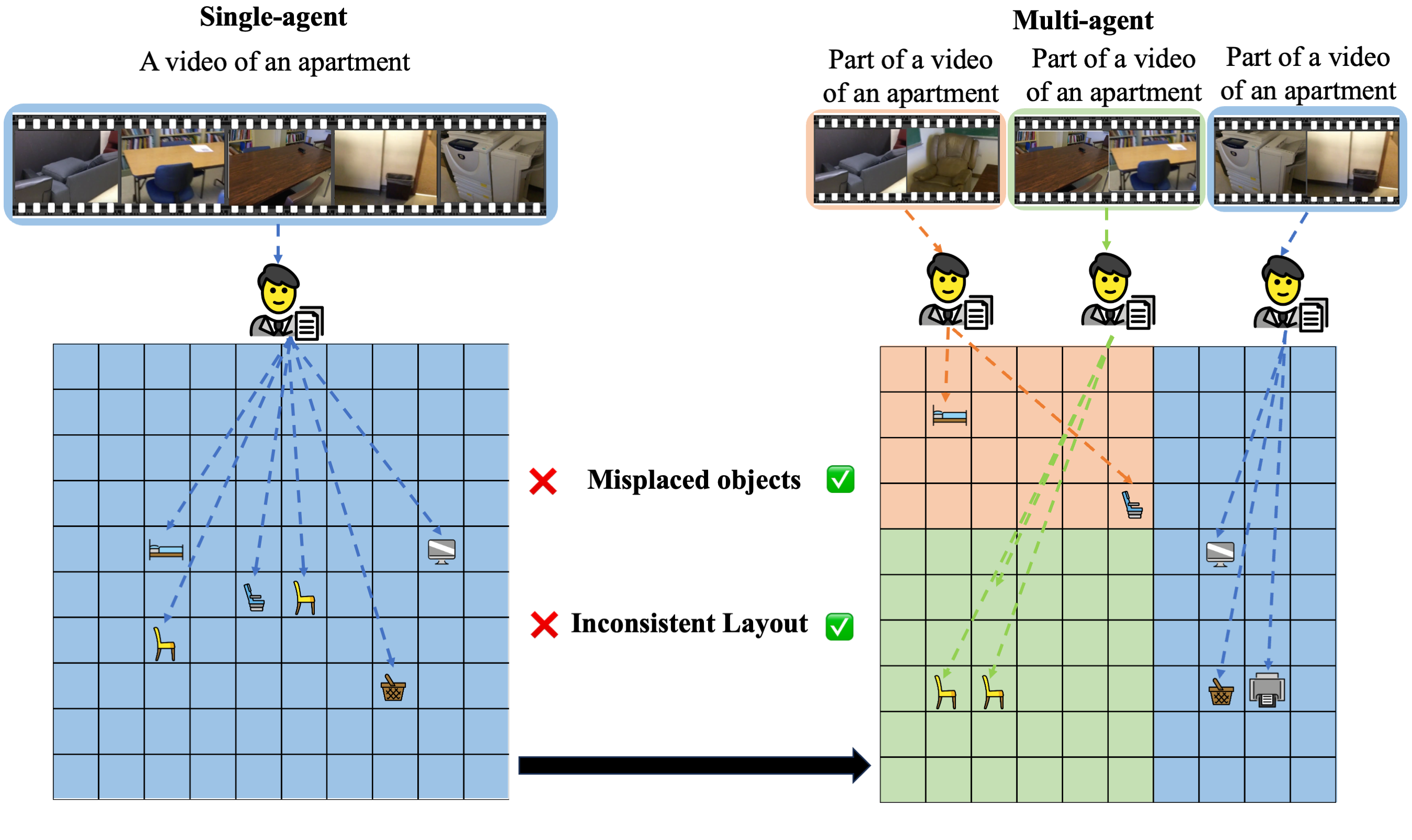}
    \caption{Teaser of our multi-agent framework for video spatial understanding.
Given a video of an apartment, a single-agent MLLM processes the entire sequence at once and is prone to missing objects, misplacing objects, and producing an inconsistent spatial layout due to context limitations. In contrast, our multi-agent framework decomposes the video into shorter segments processed by multiple agents, whose local observations are then integrated into a more complete, accurate, and consistent global cognitive map.}
    \label{fig:teaser}
\end{figure}

Spatial intelligence is a fundamental capability for multimodal large language models (MLLMs), enabling them to perceive, organize, and reason about the physical world from visual observations.~\cite{chen2024spatialvlm, yang2025thinking} 
Although recent MLLMs have shown strong performance on image and video understanding, spatial reasoning over long-horizon videos remains challenging. 
Unlike static image understanding, video-based spatial reasoning requires models to integrate observations across time, remember previously seen objects, infer their spatial relationships, and maintain a coherent representation of the surrounding environment.
In this context, cognitive maps, as introduced in~\cite{yang2025thinking}, offer a lightweight yet effective intermediate representation for organizing multi-frame observations into structured spatial layouts, thereby supporting persistent and coherent spatial reasoning.
However, the limited context length of current MLLMs makes dense long-video processing impractical.~\cite{shen2024longvu,shu2025video}
Consequently, a single MLLM understanding a long video's spatial understanding is prone to object mislocalization and inconsistent layout construction.
% Therefore, existing systems commonly rely on uniform frame sampling or keyframe selection before feeding videos into the model.~\cite{weng2024longvlm,song2024moviechat}
% This introduces an inherent trade-off: dense sampling retains more visual evidence but quickly exhausts the context budget, while sparse sampling reduces the input length but may discard critical observations needed for accurate spatial reasoning. 
% 

To address the context limitation of MLLMs, prior studies have explored several directions. 
Long-context MLLMs extend the input window to accommodate more frames, allowing models to process longer videos directly.~\cite{shu2025video}
Other approaches introduce external memory modules to store and retrieve past observations.~\cite{song2024moviechat, he2024ma}
These methods have shown promise in preserving long-range information and supporting more consistent reasoning. 
However, many of them require modifying model architectures, adding task-specific memory components, or finetuning the underlying MLLM. This limits their applicability to existing pretrained models and increases the cost of deployment, especially when the goal is to leverage powerful off-the-shelf MLLMs without additional training.

To address these challenges, we first observe that the main bottleneck lies in relying on a single MLLM to both memorize long-horizon visual observations and reason over their spatial relationships within a limited context window. 
This often leads to misplaced objects and inconsistent layout. 
Meanwhile, the cognitive map representation introduced in~\cite{yang2025thinking} offers a concise and lightweight way to encode spatial information, where objects and their relative positions are organized into a structured layout. 
However, constructing such a map from long videos remains difficult for a single model, since local observations may be scattered across distant frames and require consistent integration over time. 
This motivates a collaborative formulation, where multiple agents focus on local segments and jointly contribute to a unified spatial memory.
In this work, we propose a plug-and-play multi-agent framework for video spatial understanding, with an overview shown in Fig.~\ref{fig:teaser}. 
It decomposes long videos into shorter segments, assigns them to multiple agents for local spatial reasoning, and coordinates their partial observations into a unified cognitive map. 
To improve reliability, we introduce atomic commits for controlled map updates and cross-agent verification for reconciling inconsistent object localization and layout estimation. 
Our framework requires no architectural modification or finetuning, and can be directly applied to arbitrary pretrained MLLMs.
% To ensure reliable map construction, we introduce cognitive map construction with atomic commits, which updates the global map through controlled local operations, reducing the risk of propagating noisy or incomplete observations. 
% We further design a cross-agent verification mechanism that compares and reconciles observations from different agents, improving robustness against inconsistent object localization and layout estimation. 
% Importantly, our framework requires no architectural modification or finetuning, and can be directly applied to arbitrary pretrained MLLMs. \textcolor{red}{according to fig 1}

Our contributions are summarized as follows:

\begin{itemize}
    \item We propose a plug-and-play multi-agent framework for video spatial understanding. Our method enables off-the-shelf MLLMs to construct cognitive maps and perform spatial reasoning in a fully training-free manner, without architectural modification or finetuning.

    \item We introduce reliable collaborative mechanisms for multi-agent spatial reasoning. Specifically, our framework combines local-global coordination for integrating segment-level observations, atomic commits for controlled cognitive map construction, and cross-agent verification for resolving conflicting observations and improving map consistency.

    \item We conduct extensive evaluations on two spatial understanding benchmarks. Experiments show that our framework consistently improves over single-agent methods, general multi-agent baselines, and other training-free alternatives, demonstrating its effectiveness for video spatial reasoning.
\end{itemize}

\section{Related Work}

\textbf{MLLMs for Long Video Understanding}
Video understanding has moved from short-clip recognition to long-horizon multimodal reasoning. Early video-language pretraining methods, such as VideoBERT~\cite{sun2019videobert} and HERO~\cite{li2020hero}, showed that transformers can align video events with text and improved temporal modeling for long video retrieval and question answering. With the rise of multimodal large language models, Flamingo~\cite{alayrac2022flamingo} and VideoChatGPT~\cite{maaz2024video} showed that LLMs with visual encoders can handle open-ended video question answering and instruction following. However, these methods often use sparse frame sampling or short context windows, so they may miss long-range temporal information.
Recent work addresses this scalability bottleneck directly. LongVILA~\cite{chen2024longvila} expands multimodal context windows to millions of tokens through sequence-parallel training, while InternVideo2.5~\cite{wang2025internvideo2} improves efficiency via adaptive hierarchical token compression. Beyond longer contexts, VideoAgent~\cite{wang2024videoagent} maintains memory over events and object states while using tool calls for clip localization, and MR. Video~\cite{pangmr} decomposes long-video reasoning into local segment encoding followed by global aggregation under a MapReduce paradigm. Collectively, these works indicate a transition from passive long-context scaling to memory-augmented and structure-aware long-horizon video reasoning.
Our work proposes a training-free, multi-agent architecture for flexible and efficient long-horizon video reasoning with a plug-and-play design.

\textbf{Memory for Spatial Reasoning}
Spatial reasoning requires agents to keep a persistent world model instead of relying on a single image. Early embodied AI systems used explicit memory for this goal. Neural Map ~\cite{parisotto2017neural} introduced a learnable spatial memory for partially observable environments and showed that external memory helps long-horizon navigation. Active Neural SLAM ~\cite{chaplot2020learning} further combined mapping, localization, and planning in one differentiable framework, showing the value of structured geometric memory. Later work added semantic information to spatial maps.  SpatialVLM ~\cite{chen2024spatialvlm} introduces large-scale supervision for metric and relational 3D reasoning, significantly improving quantitative spatial question answering. Spatial-MLLM ~\cite{wu2025spatial} separates semantic and geometric encoding through a dual-encoder design, combining 2D perception with dedicated spatial features. Beyond perception, ConceptGraphs ~\cite{gu2024conceptgraphs} constructs open-vocabulary 3D scene graphs from multi-view observations, offering reusable symbolic world representations for planning. 3D-Mem ~\cite{yang20253d} further models explored and unexplored regions through snapshot memories, enabling reasoning over both observed space and future exploration targets. 
Unlike dense metric maps or full 3D scene graphs, our framework adopts cognitive maps as lightweight, query-relevant spatial memories: each agent builds a local map from its assigned video segment, and these maps are collaboratively aligned, verified, and integrated into a unified global cognitive map for spatial reasoning.

\textbf{Multi-Agent Collaboration}
Recent studies explored whether complex reasoning can be improved by using multiple interacting agents. An early related idea is self-ensemble reasoning, where different solution paths are generated and combined. For example, Self-Consistency ~\cite{wang2022self} showed that sampling multiple chains of thought and selecting the consensus answer can improve reasoning performance. This suggests that diversity and aggregation are useful even in a single model. Later, tool-augmented reasoning expanded this idea. ReAct ~\cite{yao2022react} combines reasoning with actions, allowing models to retrieve information and update plans during problem solving. Soon after, multi-agent frameworks such as CAMEL ~\cite{li2023camel} and AutoGen ~\cite{wu2024autogen} showed that agents with different roles can collaborate through dialogue and solve complex tasks more effectively than a single model.
Recent work extends collaborative reasoning to multimodal and embodied settings. Multiagent Debate ~\cite{du2024improving} formalizes iterative argument exchange among multiple LLM instances, improving factuality and reasoning robustness. LVAgent ~\cite{chen2025lvagent} proposes a multi-MLLM collaboration pipeline in which agents specialize in selection, perception, action, and reflection. LongVideoAgent ~\cite{liu2025longvideoagent} applies this principle to long-video understanding, where a controller coordinates localization and visual agents through multi-round cooperation optimized with reinforcement learning. These results suggest that as tasks grow in temporal length and perceptual complexity, modular collaboration among specialized agents may offer a scalable alternative to continually increasing model size.
Our work applies multi-agent collaboration to video spatial understanding by assigning heterogeneous agents complementary roles in cognitive map construction, including local observation, global integration, and cross-agent verification.

\section{Methodology}

\begin{figure}[t!]
    \centering
    \includegraphics[width=\linewidth]{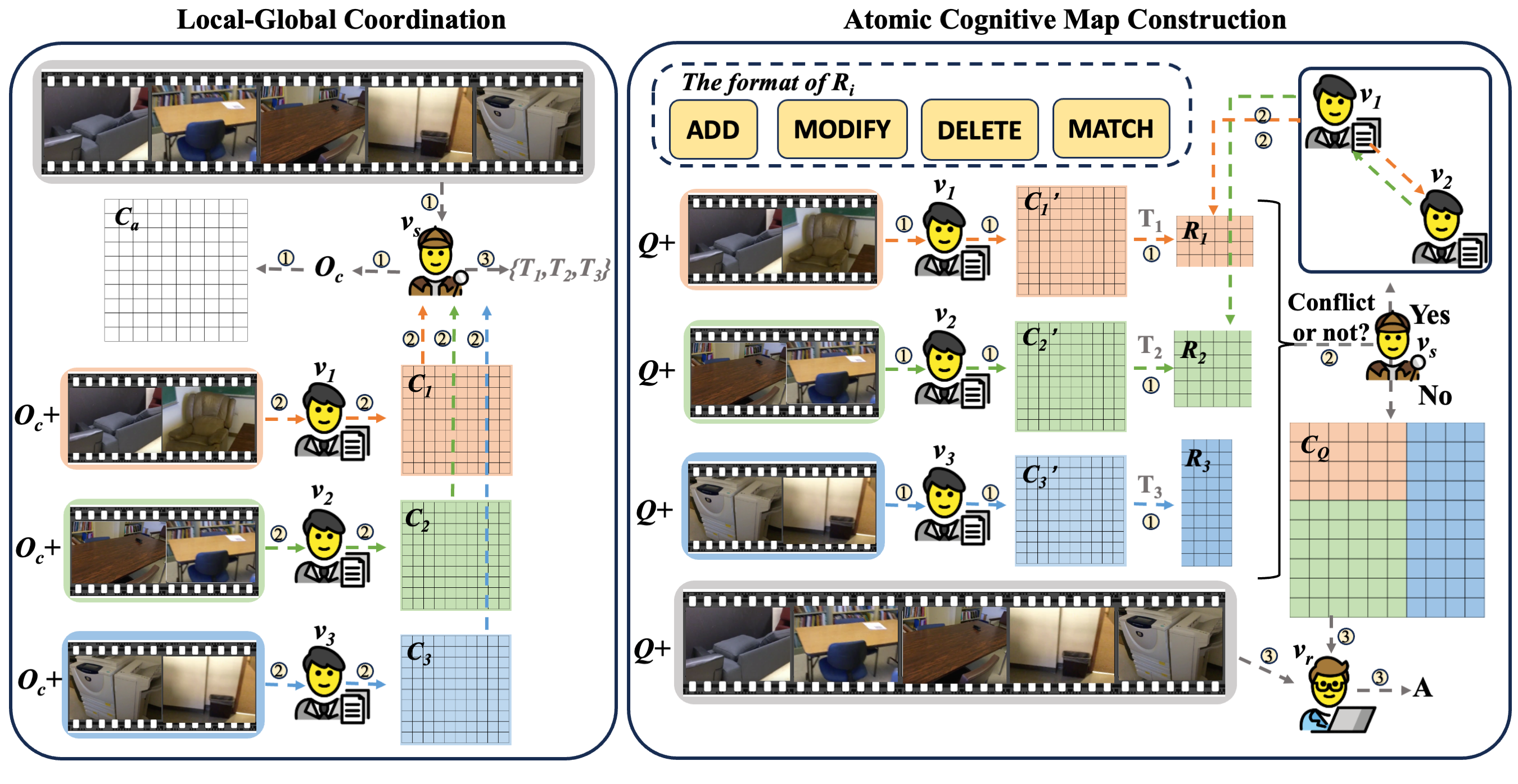}
   \caption{
Overview of our multi-agent framework for video spatial understanding.
Our method consists of two stages.
First, during \emph{local-global coordination}, a summary agent $v_s$ identifies a set of persistent objects $O_c$ that appear throughout the video and constructs a global anchor cognitive map $C_a$ over these objects.
Each local agent $v_i$ then processes an assigned video segment with respect to the shared anchor object list $O_c$ and builds a local cognitive map $C_i$.
The local maps $\{C_i\}$ are returned to the summary agent $v_s$, which estimates coordination transformations $\{T_i\}$ to align each local coordinate system with the global anchor map.
Second, during \emph{atomic cognitive map construction}, each local agent constructs a query-specific cognitive map $C'_i$ according to the spatial question $Q$.
Based on $C'_i$ and its corresponding transformation $T_i$, the local agent submits an update request $R_i$ to the summary agent $v_s$, which incrementally constructs the final cognitive map $C_Q$.
When conflicting observations arise across agents, an inter-agent debate mechanism verifies and updates inconsistent local maps before integration.
Finally, the completed map $C_Q$ is passed to a reducer agent $v_4$ to produce the answer $A$.
}
    \label{fig:design}
\end{figure}

Given a video $V$ and a spatial understanding question $Q$, our goal is to construct a query-specific cognitive map $C_Q$ for answering $Q$. 
Instead of processing the entire video with a single MLLM, our framework decomposes the video among multiple agents and coordinates their local spatial observations within a shared global reference frame. 
As shown in Figure~\ref{fig:design}, the framework first performs \emph{local-global coordination} to align local anchor maps with a global anchor cognitive map, and then conducts \emph{cognitive map construction} to integrate query-relevant local maps into $C_Q$. 
An inter-agent verification mechanism is further used to resolve inconsistent observations before integration. 
We detail these components in following sections, respectively.

\vspace{-5mm}
% \subsection{Local-Global Coordination}

\subsection{Local-Global Coordination}
\vspace{-2mm}
A key challenge in multi-agent long-video spatial understanding is aligning observations from different segments and viewpoints into a shared spatial reference. 
Existing cognitive map based approaches typically construct maps using objects directly relevant to the question and then reason over the resulting map. 
While effective for short or well-observed videos, this strategy becomes unreliable in long-horizon videos. 
Question-relevant objects may appear only briefly, be partially occluded, or be observed from limited viewpoints. 
Using such transient or ambiguous objects as spatial anchors can lead to inaccurate alignment, unstable coordinate estimation, and inconsistent layouts across video segments. 
For example, an object that appears in only one short clip may provide useful evidence for answering the question, but it is insufficient to define a stable spatial reference for the entire environment.

To address this issue, we decouple spatial anchoring from query-specific reasoning. 
Instead of directly anchoring the cognitive map on question-relevant objects, we first identify persistent objects that appear across the video and use them to establish a global spatial reference. 
These persistent objects serve as shared anchors for coordinating local agents, allowing subsequent query-specific observations to be aligned under a common coordinate system.

Formally, given a video $V$ and a spatial question $Q$, we divide the video into $N$ segments $\{V_i\}_{i=1}^{N}$. 
A summary agent $v_s$ observes the video at a coarse level and extracts a set of persistent anchor objects $O_c = \{o_1, o_2, \ldots, o_M\}$, where each object in $O_c$ is repeatedly observed across the video and can serve as a stable spatial reference. 
We represent each cognitive map as a continuous $10 \times 10$ spatial grid, where object positions are defined over $[0,10]^2$ rather than restricted to discrete cells.
Based on the anchor objects, the summary agent constructs a global anchor cognitive map
\begin{equation}
C_a = \{(o_j, \mathbf{p}^{a}_j, \mathbf{d}^{a}_j) \mid o_j \in O_c\},
\end{equation}
where $\mathbf{p}^{a}_j \in [0,10]^2$ denotes the continuous grid position of anchor object $o_j$, and $\mathbf{d}^{a}_j$ denotes its brief description in the global coordinate system.
Each local agent $v_i$ is assigned a video segment $V_i$ and constructs a local anchor map, also represented on a continuous $10 \times 10$ grid, with respect to the same anchor object list:
\begin{equation}
C_i = \{(o_j, \mathbf{p}^{i}_j, \mathbf{d}^{i}_j) \mid o_j \in O_c,\; o_j \text{ is observed in } V_i\},
\end{equation}
where $\mathbf{p}^{i}_j \in [0,10]^2$ and $\mathbf{d}^{i}_j$ denote the continuous grid position and description of anchor object $o_j$ estimated from segment $V_i$. 
Since different local agents may observe the anchors from different viewpoints or coordinate systems, the local maps $\{C_i\}$ cannot be directly merged. 
To align them, the summary agent estimates a coordination transformation $T_i$ for each local agent by matching the shared anchor objects between $C_i$ and the global anchor map $C_a$:
\begin{equation}
T_i = v_s(C_a, C_i), \quad 
T_i = (\theta_i, \mathbf{b}^{\mathrm{tl}}_i, \mathbf{b}^{\mathrm{br}}_i),
\end{equation}
where $\theta_i$ denotes the rotation angle used to align the orientation of local map $C_i$ with the global coordinate system, and $\mathbf{b}^{\mathrm{tl}}_i, \mathbf{b}^{\mathrm{br}}_i \in [0,10]^2$ denote the top-left and bottom-right coordinates of the region where the transformed local map should be placed in the global $10 \times 10$ grid. 
Thus, $T_i$ specifies both the orientation and spatial placement of local map $C_i$ within the global cognitive map $C_a$.
In this way, local-global coordination provides a stable spatial calibration before query-specific cognitive map construction, ensuring that local observations from different video segments can be consistently integrated into a unified global map.

\subsection{Atomic Cognitive Map Construction}

After local-global coordination, the next step is to construct a query-specific cognitive map containing the spatial evidence needed to answer the question. 
A straightforward solution is to ask a single MLLM to process the entire video and directly build a global map, but this is constrained by limited context length and often requires aggressive frame sampling, which may discard short-lived yet question-critical objects. 
Moreover, end-to-end reasoning over long videos can dilute local spatial details, leading to object omission, inaccurate localization, and inconsistent spatial relations. 
To preserve fine-grained evidence, we let each local agent focus on a shorter segment and construct a local query-relevant cognitive map, which is then aligned to the global reference frame using the transformations $\{T_i\}$ estimated during local-global coordination. 
Since directly merging local maps may still introduce duplicated objects, noisy observations, or conflicting spatial estimates, we formulate map integration as an atomic update process, where each local agent submits structured update requests to the summary agent rather than directly modifying the global map.

Formally, given the spatial question $Q$, each local agent $v_i$ receives its assigned video segment $V_i$ and the corresponding coordination transformation $T_i$. 
The agent first identifies query-relevant objects $O_i^Q = \{o \mid o \text{ is relevant to } Q\}$, including objects explicitly mentioned in the question and contextual objects necessary for inferring spatial relationships. 
It then constructs a query-specific local cognitive map
\begin{equation}
C'_i = v_i(V_i, Q, T_i)
= \{(o, \mathbf{p}^{i}_o, \mathbf{d}^{i}_o) \mid o \in O_i^Q\},
\end{equation}
where $\mathbf{p}^{i}_o \in [0,10]^2$ denotes the continuous position of object $o$ in the local $10 \times 10$ grid, and $\mathbf{d}^{i}_o$ denotes its brief visual description. 
Although $C'_i$ is constructed from local observations, the agent is provided with $T_i$ so that it can reason about how its local map should be transformed and placed in the global coordinate system.
Based on the local map $C'_i$ and transformation $T_i$, the local agent further generates an atomic update request:
\begin{equation}
R_i = v_i(C'_i, T_i)
= \{(a_k, o_k, \tilde{\mathbf{p}}_k, \tilde{\mathbf{d}}_k)\}_{k=1}^{K_i},
\quad
a_k \in \{\texttt{ADD}, \texttt{MODIFY}, \texttt{DELETE}, \texttt{MATCH}\}.
\end{equation}
Here, $R_i$ is expressed with respect to the global $10 \times 10$ coordinate system specified by $T_i$. 
Each operation specifies an atomic edit to the global cognitive map. 
\texttt{ADD} inserts a newly observed object into $C_Q$; \texttt{MODIFY} updates the position or description of an existing object; \texttt{DELETE} removes noisy, duplicated, or irrelevant entries; and \texttt{MATCH} associates a local object with an existing global object to avoid duplicate entities. 
The transformed position $\tilde{\mathbf{p}}_k$ and description $\tilde{\mathbf{d}}_k$ are derived from the local map $C'_i$ under the transformation $T_i$.
The summary agent $v_s$ serves as the map maintainer and decides whether each request can be committed to the global map. 
If the request is consistent with the current map, it is atomically applied:
\begin{equation}
C_Q \leftarrow \mathrm{Commit}(C_Q, R_i).
\end{equation}
If conflicts are detected, such as inconsistent object localization, duplicated entities, or contradictory descriptions, the request is not immediately committed. 
Instead, it is returned to the corresponding local agent for revision and may trigger cross-agent verification with other agents that observed the conflicting objects. 
This design prevents noisy or incomplete local observations from being directly propagated into the global cognitive map.
After all verified requests are committed, we obtain the final query-specific cognitive map $C_Q$. 
A reducer agent $v_r$ then produces the answer based on the verified cognitive map, the original video, and the spatial question:
\begin{equation}
A = v_r(C_Q, V, Q).
\end{equation}
The cognitive map provides an explicit and globally aligned spatial memory, while the original video offers complementary visual evidence for fine-grained appearance or temporal cues. 
Together, they allow the reducer agent to answer the question using both structured spatial reasoning and the original visual context.

\subsection{Iterative Cross-Agent Verification}
\label{sec:debate}

Although local-global coordination provides a shared coordinate system and atomic requests prevent uncontrolled map updates, local observations may still be noisy or inconsistent. 
Each local agent only observes a short video segment, so its request $R_i$ may contain uncertain object locations, incomplete descriptions, duplicated entities, or incorrect matches with objects already in the global map. 
If such requests are directly committed, local errors can contaminate the query-specific cognitive map $C_Q$ and accumulate during subsequent reasoning. 
To improve reliability, we introduce an iterative cross-agent verification mechanism that checks each request before committing it to the global map.

When a local agent $v_i$ submits an update request $R_i$, the summary agent $v_s$ verifies it against the current global cognitive map $C_Q$ and previously committed observations. 
The verification process is denoted as
\begin{equation}
(f_i, D_i, v_j) = v_s(R_i, C_Q),
\end{equation}
where $f_i \in \{0,1\}$ indicates whether a conflict exists, $D_i$ describes the conflicting object locations, descriptions, or entity matches, and $v_j$ denotes the agent associated with the conflicting observation. 
If $f_i=0$, the request is considered consistent and is atomically committed:
\begin{equation}
C_Q \leftarrow \mathrm{Commit}(C_Q, R_i).
\end{equation}

If $f_i=1$, the request is not committed immediately. 
Instead, the summary agent returns the conflict description $D_i$ to the proposing agent $v_i$ and asks it to verify the conflict with the corresponding agent $v_j$. 
The two agents compare their local observations, including the relevant video segments, object descriptions, and estimated positions, and then revise the original request:
\begin{equation}
\widetilde{R}_i = \mathrm{Debate}(v_i, v_j, R_i, D_i).
\end{equation}
The revised request $\widetilde{R}_i$ is resubmitted to the summary agent for verification. 
This process is repeated until the request is accepted or no further reliable revision can be made:
\begin{equation}
C_Q \leftarrow \mathrm{Commit}(C_Q, \widetilde{R}_i)
\quad \text{if } v_s(\widetilde{R}_i, C_Q) \text{ reports no conflict}.
\end{equation}

Through this verify--debate--commit process, the global cognitive map is updated only with requests that are consistent with both local evidence and existing global memory. 
The mechanism reduces duplicated objects, unstable localization, and contradictory descriptions across agents, while preserving the fine-grained spatial evidence captured from individual video segments.

\section{Experiment}

\subsection{Experimental Setup}

We evaluate our method on two video spatial understanding benchmarks: 
\textbf{VSI-Bench}~\cite{yang2025thinking} and \textbf{MMSI-Bench}~\cite{lin2025mmsi}. 
These benchmarks cover diverse spatial reasoning scenarios, including object localization, spatial relation understanding, temporal integration, and long-horizon environment-level reasoning. 
They require models to aggregate visual evidence across video frames and maintain a coherent representation of the observed space. For fair comparison, all methods are evaluated under identical experimental settings on the same video samples, using subsets of 500 QA pairs from VSI-Bench and 500 QA pairs from MMSI-Bench due to limited resources.

To examine the generality of our plug-and-play framework, we evaluate it with both closed-source and open-source MLLMs. 
For closed-source models, we use representative proprietary models including Gemini-2.5-Flash~\cite{gemini2025} and GPT-5-mini~\cite{singh2025openai}. 
For open-source models, we adopt Qwen2.5-VL-32B~\cite{qwen} as the backbone. 
For each backbone, we compare our multi-agent framework with the corresponding single-agent baseline under the same prompting protocol and input-frame budget. 
This setup isolates the effect of our multi-agent cognitive map construction from gains introduced by stronger backbone models or additional visual inputs.

\begin{table*}[h!]
\centering
\caption{
Main results on video spatial understanding benchmarks with different base VLMs.
Across GPT-, Gemini-, and Qwen-VL-based backbones, our training-free framework consistently outperforms single-agent prompting and representative training-free agentic video understanding baselines.
These results demonstrate the effectiveness of combining agentic collaboration with explicit cognitive map construction for video spatial understanding.
}
\label{tab:main_results_all}

\begin{subtable}{\textwidth}
\centering
\caption{GPT-based backbone.}
\label{tab:gpt_main}
\resizebox{\textwidth}{!}{
\begin{tabular}{lccccc}
\toprule
\textbf{Method}
& \textbf{Training-free}
& \textbf{Agentic Collaboration}
& \textbf{CogMap}
& \textbf{VSI-Bench}
& \textbf{MMSI-Bench} \\
\midrule
Single-Agent Baseline
& \cmark
& \xmark
& \cmark
& 39.7 
& 35.0 \\

TraveLER~\cite{shang2024traveler}
& \cmark
& \cmark
& \xmark
& 39.3
& 34.7 \\

VideoAgent~\cite{wang2024videoagent}
& \cmark
& \xmark
& \xmark
& 38.4
& 33.8 \\

VCA~\cite{yang2025vca}
& \cmark
& \xmark
& \xmark
& 38.9
& 34.2 \\

\midrule
\textbf{Ours}
& \cmark
& \cmark
& \cmark
& \textbf{43.0}
& \textbf{39.1} \\
\bottomrule
\end{tabular}
}
\end{subtable}

\vspace{0.8em}

\begin{subtable}{\textwidth}
\centering
\caption{Gemini-based backbone.}
\label{tab:gemini_main}
\resizebox{\textwidth}{!}{
\begin{tabular}{lccccc}
\toprule
\textbf{Method}
& \textbf{Training-free}
& \textbf{Agentic Collaboration}
& \textbf{CogMap}
& \textbf{VSI-Bench}
& \textbf{MMSI-Bench} \\
\midrule
Single-Agent Baseline
& \cmark
& \xmark
& \cmark
& 44.9
& 36.6 \\

TraveLER~\cite{shang2024traveler}
& \cmark
& \cmark
& \xmark
& 44.1
& 36.0 \\

VideoAgent~\cite{wang2024videoagent}
& \cmark
& \xmark
& \xmark
& 43.3
& 35.4 \\

VCA~\cite{yang2025vca}
& \cmark
& \xmark
& \xmark
& 43.6
& 35.8 \\

\midrule
\textbf{Ours}
& \cmark
& \cmark
& \cmark
& \textbf{49.7}
& \textbf{41.0} \\
\bottomrule
\end{tabular}
}
\end{subtable}

\vspace{0.8em}

\begin{subtable}{\textwidth}
\centering
\caption{Qwen-VL-based backbone.}
\label{tab:qwen_main}
\resizebox{\textwidth}{!}{
\begin{tabular}{lccccc}
\toprule
\textbf{Method}
& \textbf{Training-free}
& \textbf{Agentic Collaboration}
& \textbf{CogMap}
& \textbf{VSI-Bench}
& \textbf{MMSI-Bench} \\
\midrule
Single-Agent Baseline
& \cmark
& \xmark
& \cmark
& 36.6 
& 32.4 \\

TraveLER~\cite{shang2024traveler}
& \cmark
& \cmark
& \xmark
& 36.2
& 31.9 \\

VideoAgent~\cite{wang2024videoagent}
& \cmark
& \xmark
& \xmark
& 35.4
& 31.2 \\

VCA~\cite{yang2025vca}
& \cmark
& \xmark
& \xmark
& 35.7
& 31.6 \\

\midrule
\textbf{Ours}
& \cmark
& \cmark
& \cmark
& \textbf{41.1}
& \textbf{35.3} \\
\bottomrule
\end{tabular}
}
\end{subtable}

\end{table*}

\vspace{2mm}
\subsection{Main Results: Comparison with Baselines}

We compare our method with several representative training-free and agentic baselines for video understanding, including a \textit{single-agent baseline}, \textit{TraveLER}~\cite{shang2024traveler}, \textit{VideoAgent}~\cite{wang2024videoagent}, and \textit{VCA}~\cite{yang2025vca}. The single-agent baseline processes the entire video with one VLM, while TraveLER, VideoAgent, and VCA use agentic planning, evidence retrieval, or autonomous video exploration for long-form video reasoning.

As shown in Table~\ref{tab:main_results_all}, our method achieves the best performance across all backbones and benchmarks. Compared with the single-agent baseline, it decomposes long videos into agent-level observations and enables collaborative reasoning, reducing the burden of long-context modeling. Compared with TraveLER~\cite{shang2024traveler}, VideoAgent~\cite{wang2024videoagent}, and VCA~\cite{yang2025vca}, our method further introduces explicit cognitive maps, which preserve and share spatial relationships across agents, leading to stronger long-range spatial understanding and more accurate multi-step reasoning.

\begin{figure}[h!]
    \centering
    \begin{subfigure}[t]{0.32\linewidth}
        \centering
        \includegraphics[width=\linewidth]{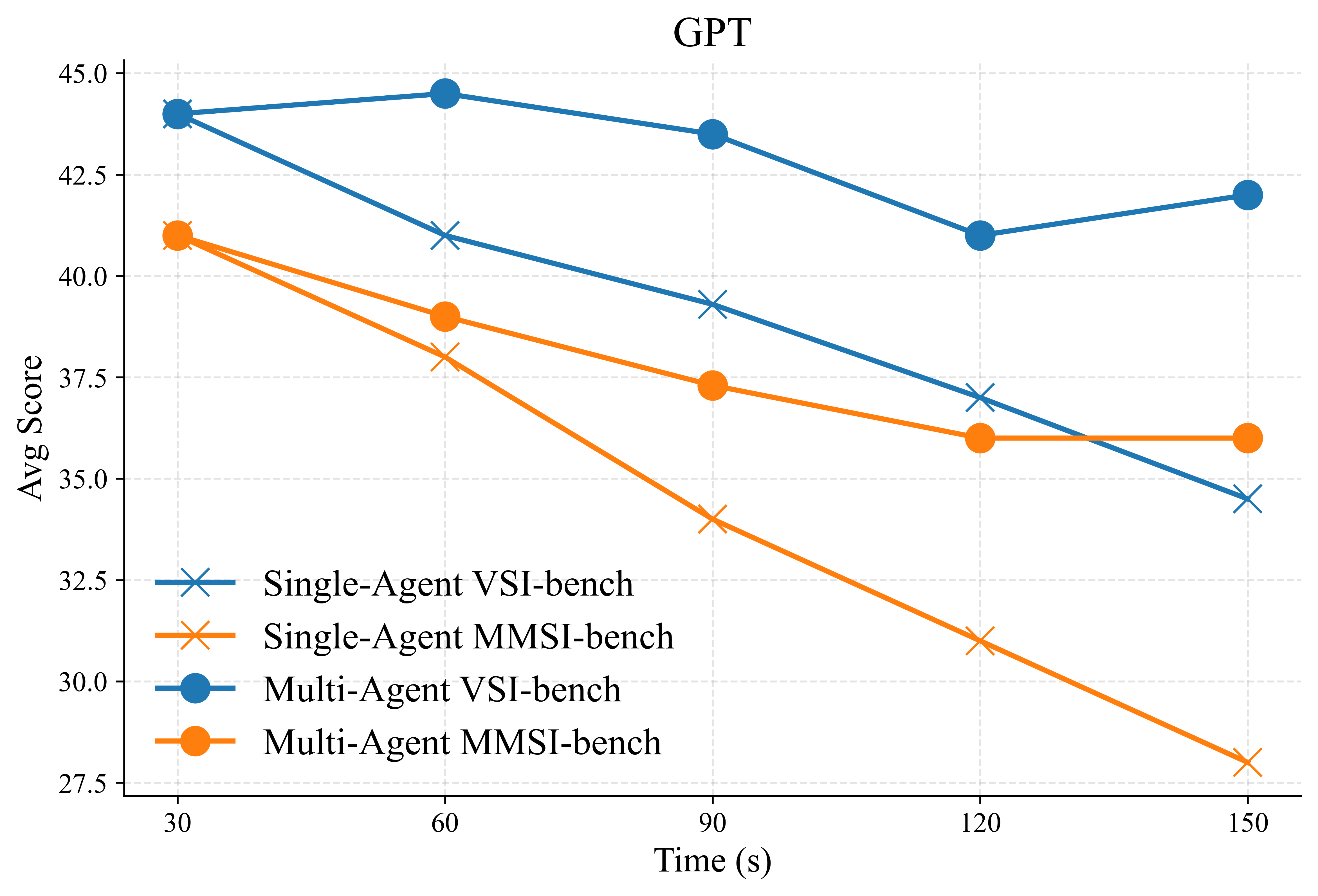}
        \caption{GPT-based backbone}
        \label{fig:length_gpt}
    \end{subfigure}
    \hfill
    \begin{subfigure}[t]{0.32\linewidth}
        \centering
        \includegraphics[width=\linewidth]{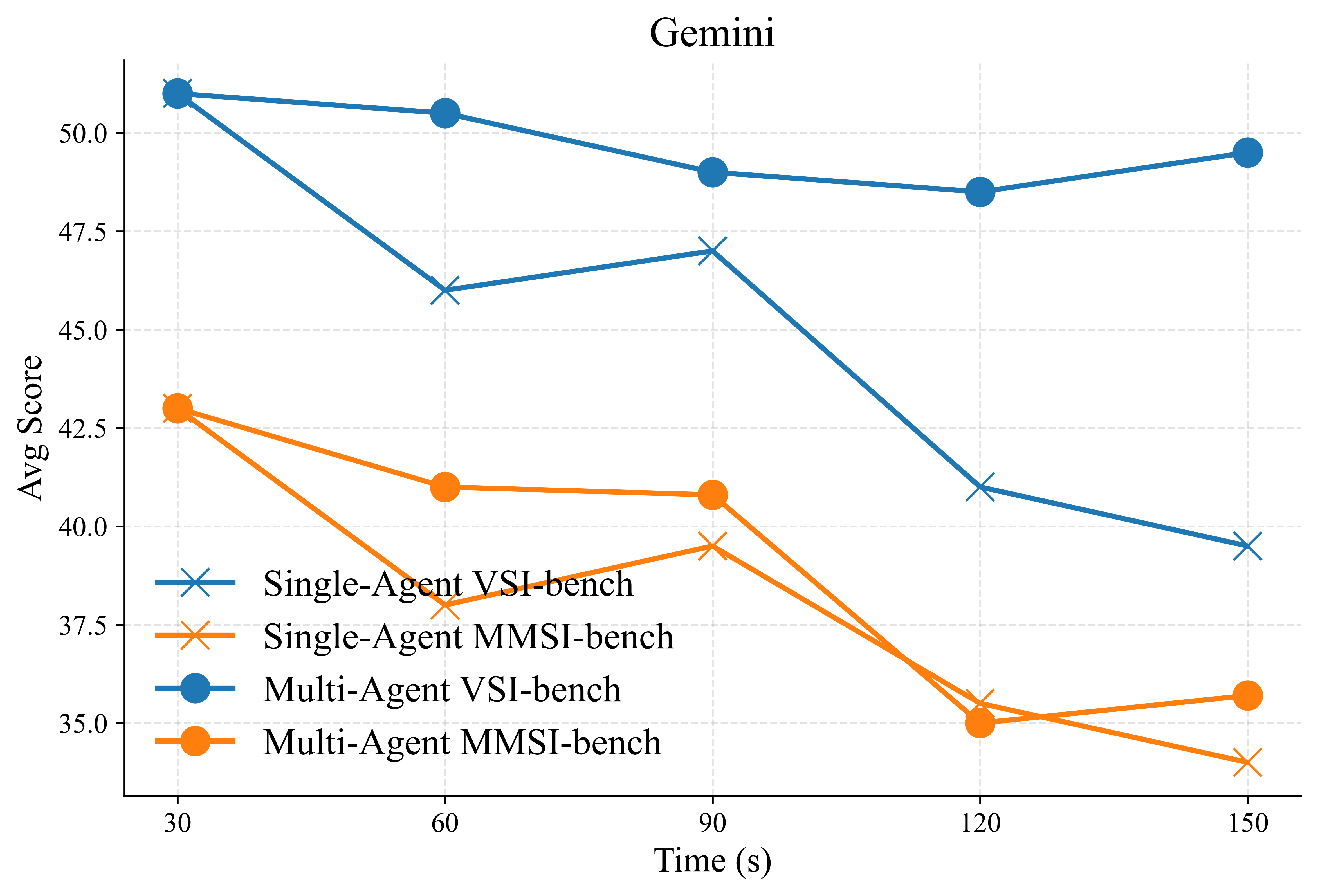}
        \caption{Gemini-based backbone}
        \label{fig:length_gemini}
    \end{subfigure}
    \hfill
    \begin{subfigure}[t]{0.32\linewidth}
        \centering
        \includegraphics[width=\linewidth]{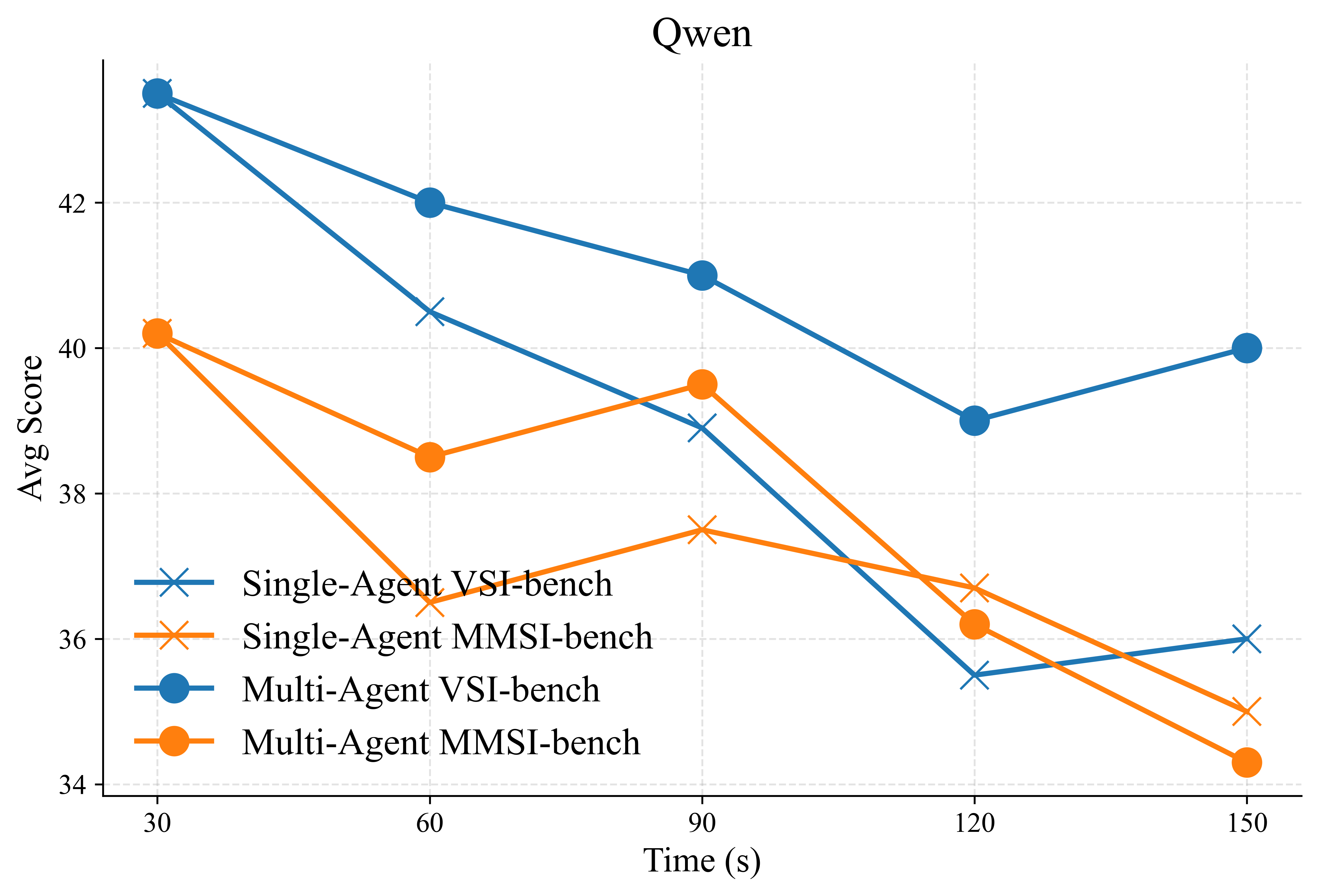}
        \caption{Qwen-VL-based backbone}
        \label{fig:length_qwen}
    \end{subfigure}
    \caption{
    Accuracy trends with respect to video length across three VLM backbones.
    Videos are grouped by duration, and we compare our method with the single-agent baseline.
    Although accuracy decreases as videos become longer, our method shows a slower performance degradation across all three backbones.
    }
    \label{fig:length_analysis}
\end{figure}

We also study the effect of video length on performance across GPT-, Gemini-, and Qwen-VL-based models. As shown in Figure~\ref{fig:length_analysis}, accuracy generally decreases as videos become longer, but our method shows a slower drop than the single-agent baseline across all backbones. This suggests that agentic decomposition and cognitive map construction help preserve task-relevant spatial information over longer temporal horizons, improving robustness on long videos.

The consistent improvements across GPT-, Gemini-, and Qwen-VL-based backbones demonstrate that the proposed framework is not tied to a specific VLM.
Instead, the gains come from the integration of agentic collaboration with explicit spatial memory.
These results verify the effectiveness of our training-free design and highlight the importance of 3D-aware cognitive map construction for video spatial understanding. 

\subsection{Case Study}

\begin{figure}[h!]
    \centering
    \includegraphics[width=0.9\linewidth]{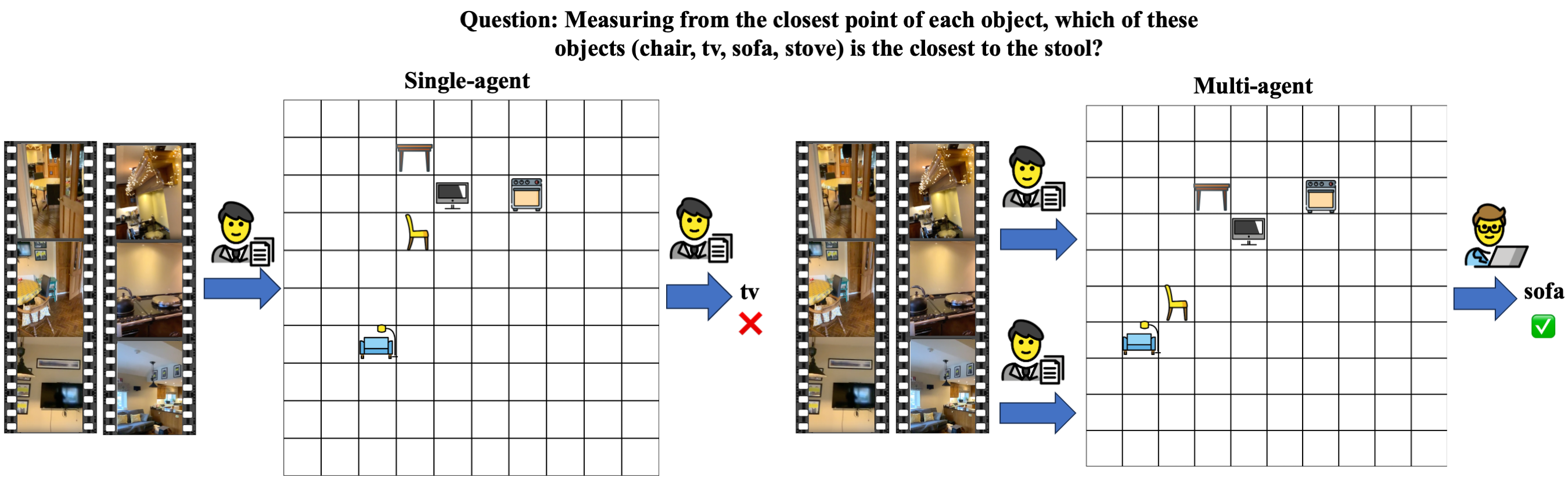}
    \caption{
    Qualitative comparison between single-agent and multi-agent cognitive map construction.
    Given the question of which object is closest to the stool, the single-agent method constructs an inaccurate spatial map and incorrectly predicts the TV.
    In contrast, CoCoSI aggregates complementary observations from multiple agents, producing a more coherent cognitive map and correctly identifying the sofa as the closest object.
    }
    \label{fig:example}
\end{figure}

As shown in Figure~\ref{fig:example}, collaborative cognitive map construction supports more reliable spatial reasoning than directly constructing a map from the entire video with a single agent. In this example, the task is to determine which object among the chair, TV, sofa, and stove is closest to the stool. The single-agent method incorrectly predicts the TV due to inaccurate spatial relationships in the constructed map. By contrast, CoCoSI decomposes the video into multiple segments, enabling different agents to capture complementary local observations. These observations are then aggregated into a more coherent cognitive map that preserves finer-grained spatial evidence. As a result, CoCoSI correctly identifies the sofa as the closest object to the stool, highlighting the benefit of multi-agent collaboration for accurate spatial reasoning.

\subsection{Ablation Study}

\begin{table*}[h!]
\centering
\caption{
Ablation study on video spatial understanding benchmarks using Gemini as the backbone.
We evaluate the contribution of three core components in our framework: local-global coordination, atomic commit, and inter-agent debate.
Each variant removes one component while keeping the remaining modules unchanged.
}
\label{tab:ablation}
\resizebox{\textwidth}{!}{
\begin{tabular}{lccccc}
\toprule
\textbf{Variant}
& \textbf{Local-Global Coord.}
& \textbf{Atomic Commit}
& \textbf{Inter-Agent Debate}
& \textbf{VSI-Bench}
% & \textbf{SPAR-Bench}
& \textbf{MMSI-Bench} \\
\midrule
w/o Local-Global Coord.
& \xmark
& \cmark
& \cmark
& 46.0 & 35.6  \\

w/o Atomic Commit
& \cmark
& \xmark
& \cmark
& 45.2 & 37.0  \\

w/o Inter-Agent Debate
& \cmark
& \cmark 
& \xmark 
& 47.8 & 39.1 \\

\midrule
\textbf{Full Model}
& \cmark
& \cmark
& \cmark
& \textbf{49.7} & \textbf{41.0}  \\
\bottomrule
\end{tabular}
}
\end{table*}

We conduct ablation studies using Gemini as the backbone to analyze the contribution of each key component in our framework.
As shown in Table~\ref{tab:ablation}, removing any component leads to a clear performance drop on both VSI-Bench and MMSI-Bench, demonstrating that the proposed modules are complementary and jointly contribute to robust video spatial understanding.
The full model achieves the best results, indicating that accurate spatial reasoning requires not only agentic decomposition, but also explicit spatial alignment, reliable memory updates, and collaborative verification.

\paragraph{Effect of Local-Global Coordination.}
Removing local-global coordination causes a clear performance drop, especially on MMSI-Bench. Without explicit alignment, agents rely mainly on textual communication to merge local observations, which is insufficient for precise cross-view spatial reasoning. In contrast, our full model aligns local maps with the global map, improving relative localization, cross-view correspondence, and long-range consistency.

\paragraph{Effect of Atomic Commit.}
Removing atomic commit leads to less constrained map updates and may introduce redundant, ambiguous, or conflicting spatial information. This variant achieves the lowest score on VSI-Bench, indicating the importance of reliable memory construction. By decomposing updates into atomic and verifiable operations, our full model improves map consistency and reduces error accumulation.

\paragraph{Effect of Inter-Agent Debate.}
Removing inter-agent debate also reduces performance, showing the benefit of collaborative verification. Without debate, uncertain or incorrect local observations may be directly propagated to the final map. Debate allows agents to compare observations, resolve conflicts, and refine spatial relations, thereby reducing hallucination and improving reasoning robustness.

Overall, these ablations confirm that all three components are necessary. Local-global coordination enables spatial alignment, atomic commit ensures structured memory updates, and inter-agent debate improves final-map correctness through collaborative verification.

\section{Conclusion}

We presented CoCoSI, a plug-and-play multi-agent framework for video spatial understanding with pretrained MLLMs. CoCoSI decomposes long videos into shorter segments and coordinates multiple agents to build a unified cognitive map, combining local-global collaboration, atomic map construction, and iterative cross-agent verification. Experiments on VSI-Bench and MMSI-Bench show that CoCoSI consistently improves spatial reasoning over single-agent prompting and training-free agentic baselines across closed- and open-source MLLMs. Future work includes extending cognitive maps to richer 3D, semantic, or temporally evolving spatial memories, and developing more efficient segment selection and agent allocation strategies.

\bibliographystyle{unsrtnat}
\bibliography{refs}

\end{document}